\documentclass[10pt,twocolumn,letterpaper]{article}

\usepackage{iccv}
\usepackage{times}
\usepackage{epsfig}
\usepackage{graphicx}
\usepackage{amsmath}
\usepackage{amssymb}
\usepackage{booktabs}
\usepackage{multirow}
\usepackage{makecell}
\usepackage{tabularx}
\usepackage[misc]{ifsym}

\newcommand\blfootnote[1]{%
  \begingroup
  \renewcommand\thefootnote{}\footnote{#1}
  \endgroup
}

\usepackage{color}
\definecolor{mycitecolor}{rgb}{0, 0.4, 0.7}
\usepackage[pagebackref=true,breaklinks=true,letterpaper=true,colorlinks,bookmarks=true,citecolor=mycitecolor]{hyperref}

\iccvfinalcopy 

\def\iccvPaperID{2397} 

\ificcvfinal\fi

\begin{document}

\title{3D Data Augmentation for Driving Scenes on Camera}

\author{
Wenwen Tong$^{1*}$\quad
Jiangwei Xie$^{1*}$\quad
Tianyu Li$^{2*}$\quad
Hanming Deng$^{1*}$
\quad 
Xiangwei Geng$^{2}$ \quad \\ 
Ruoyi Zhou$^{1}$\quad
Dingchen Yang$^{2}$ \quad
Bo Dai$^{2}$ \quad
Lewei Lu$^{1}$ \quad
Hongyang Li$^{2}$\textsuperscript{\Letter}
\\
[2mm]
$^{1}$SenseTime Research \quad
$^{2}$Shanghai AI Laboratory \\
}

\maketitle
\ificcvfinal\thispagestyle{empty}\fi

\begin{abstract}
Driving scenes are extremely diverse and complicated that it is impossible to collect all 
cases
with 
human effort alone. 
While data augmentation is an effective technique to enrich the training data, existing methods for camera data in autonomous driving applications are confined to the 2D image plane, which may not optimally increase data diversity in 3D real-world scenarios.
To this end, 
we propose a 3D data augmentation approach termed Drive-3DAug, aiming at
augmenting the driving scenes on camera in the 3D space.
We first utilize Neural Radiance Field (NeRF) to reconstruct the 3D models of background and foreground objects. Then, augmented driving scenes can be obtained by placing the 3D objects with adapted location and orientation at the pre-defined valid region of backgrounds. 
As such, 
the training database could be effectively scaled up.
However, the 3D object modeling is constrained to the image quality and the limited viewpoints.
To overcome these problems,
we modify the original NeRF by introducing a geometric rectified loss and a symmetric-aware training strategy.
We evaluate our method for the camera-only monocular 3D detection task on the Waymo and nuScences datasets. The proposed data augmentation approach contributes to a gain of $1.7\%$ and $1.4\%$ 
in terms of detection accuracy, on Waymo and nuScences respectively.
Furthermore, the constructed 3D models serve as digital driving assets and could be recycled for different detectors or other 3D perception tasks. 

\end{abstract}

\blfootnote{*: Equal contribution.}
\blfootnote{\Letter: Corresponding author at
\texttt{lihongyang@pjlab.org.cn}.}

\section{Introduction}
3D perception system, particularly 3D object detection, plays a vital role for autonomous driving. Despite recent progress \cite{brazil2019m3d, li2019gs3d, xu2018multi, li2020rtm3d,  mousavian20173d, qin2019monogrnet,  qin2019triangulation,  shi2021geometry, zhang2021objects, zhang2021learning, chen20153d, park2021pseudo, wang2021fcos3d, wang2022detr3d, weng2019monocular}, the current perception system still suffers from the hard case challenge due to the long-tail driving scenes on the road, \textit{e.g.} trucks with diversified pose on the road. 
To overcome this challenge, data augmentation has been proven to be an effective technique to enrich training data. For LiDAR-based 3D perception, different data augmentation methods \cite{reuse2021ambiguity,fang2021lidar,abs-2004-01643} have made great achievements by generating new drive scenes. 
However, it is still under-explored how to augment the driving scenes for camera-based 3D perception with data augmentation.

\begin{figure}
  \centering
  \includegraphics[width=.47\textwidth]{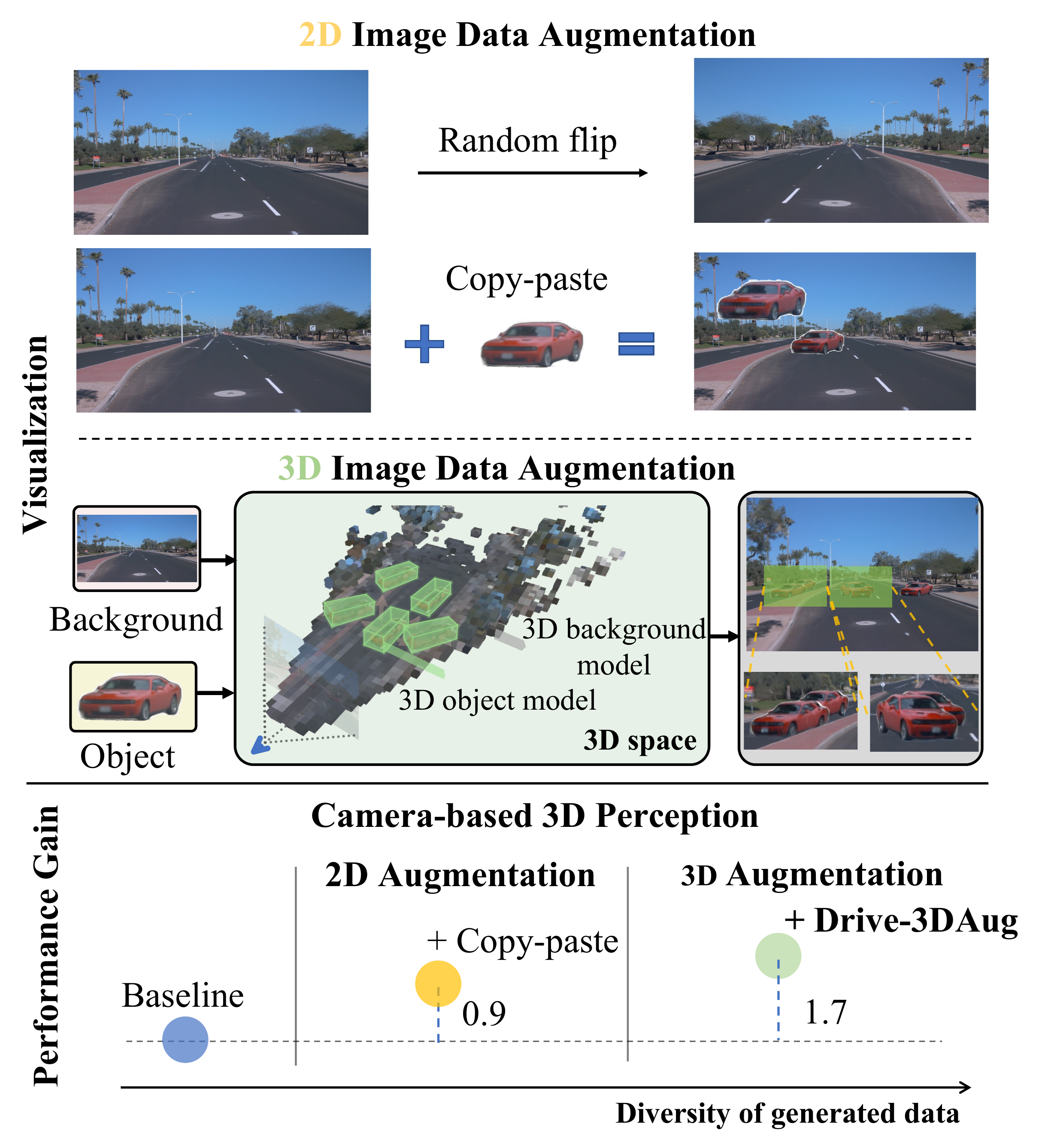}
    \caption{\textbf{Visualization of different data augmentation techniques and their \emph{performance gain} of LET-AP \cite{hung2022let3dap} for monocular 3D detection on Waymo \cite{sun2020scalability}} with 100 scenes augmented. Compared with previous 2D image data augmentations, our Drive-3DAug method, modifies the driving scenes in the 3D space. This can generate more diverse driving scenes and contribute larger performance gain to camera-based 3D perception tasks. Baseline is the FCOS3D \cite{wang2021fcos3d} method.
  }
  \label{fig:motivation}
\end{figure}

As illustrated in Figure \ref{fig:motivation}, existing image data augmentation approaches are mostly restricted to the 2D image plane, such as image transformations \cite{autoaug} and copy-paste \cite{copypaste, lian2022exploring}. These techniques face challenges in changing the view of the components in the scene, such as rotating objects within the image, thereby limiting the diversity of generated driving scenes. By contrast, the data augmentation approaches \cite{fang2021lidar,abs-2004-01643,abs-2208-00223} for point clouds are applied in the 3D space, offering more degrees of freedom to change the driving scenes. Although existing data augmentation approaches for image data have achieved some performance gains for the 3D detection task, such gains are still limited compared with the improvement brought by the 3D data augmentation approaches \cite{reuse2021ambiguity} for the LiDAR-based 3D perception tasks. 
This means the diversity of generated scenes is essential to improve the performance of 3D perception tasks. The manner of augmenting data in the 3D space is an effective way to create diverse scenes.
One might argue that simulator \cite{carla,ShahDLK17} is a powerful tool to generate synthetic 3D imagery to supplement the database. However, the sim2real bottleneck is a long-standing issue to be resolved.


In this work, we have pioneered research into 3D data augmentation for camera-based 3D percpetion for in autonomous driving. 
To implement 3D data augmentation for image data, we need to convert the scenes to the 3D space. This is because manipulating objects on 2D image plane satisfying 3D-imaging constraint, such as rotating pixels of objects, will generate flawed images.
One desirable solution is using Neural Radiance Field (NeRF)~\cite{mildenhall2020nerf,park2021nerfies,yu2022plenoxels} to reconstruct the 3D models of background and foreground objects obtained by decomposing the scene, then we can compose them for data augmentation.
However, it is hard to achieve perfect decomposition without ground truth to extract objects. Pixels of the object will be mixed with the background pixels near the object edge.
In addition, the limited viewpoints of objects in driving scenes make it difficult to apply NeRF for generating objects of novel views by large rotation, further limiting the diversity of generated scenes.
To this end, we present a novel 3D data augmentation approach for 2D images, named Drive-3DAug. Our approach has two stages. The first stage is to build the 3D models. We decompose the driving scenes into multiple backgrounds and foreground objects and use voxel-based NeRF \cite{sun2021direct} to turn them into 3D models.
%
To overcome the difficulties of reconstructing the driving scenes, 
we propose a geometric rectified loss to weaken the effect of noisy edges
and a symmetric-aware training strategy for objects with symmetry, such as vehicles, 
to broaden the available rendering viewpoints.
%
After the first stage, we re-compose the 3D models of backgrounds and foreground objects to create new driving scenes. Considering the physical constraints in the real world, we adopt a strategy to identify the valid region of the backgrounds. The augmented images are then generated by rendering the constructed scenes and can be further applied to the training of 3D perception tasks. Compared with existing approaches\cite{lian2022exploring}, our method contributes to a larger performance gain of the 3D detection task, as described in Figure \ref{fig:motivation}.

We evaluate Drive-3DAug for the monocular 3D detection task, as it is one of the most important 3D perception tasks in autonomous driving, on the Waymo~\cite{sun2020scalability} and nuScenes~\cite{caesar2020nuscenes} datasets with different detectors. 
Our method is able to achieve $1.7\%$ improvement of performance on Waymo, especially $2.2\%$ for vehicles with rare orientations, and $1.4\%$ on nuScenes on their corresponding metrics. 
Moreover, once the 3D models of the backgrounds and foreground objects are reconstructed, they can serve as a digital driving asset, which can be used repeatedly for different tasks.

To sum up, our contributions are three-fold:
\begin{itemize}
    \item We have pioneered research into the 3D data augmentation problem for camera-based 3D perception in autonomous driving.
    \item We propose a 3D data augmentation approach based on NeRF including improvements by a geometrically rectified loss and a symmetric-aware training strategy to generate more natural and diverse images.
    \item We evaluate our method on the Waymo and nuScenes dataset, demonstrating that it can improve the performance of 3D object detection, and the 3D models can be further used as digital driving assets.
\end{itemize}

\begin{figure*}[t]
  \centering
  \includegraphics[width=.99\textwidth]{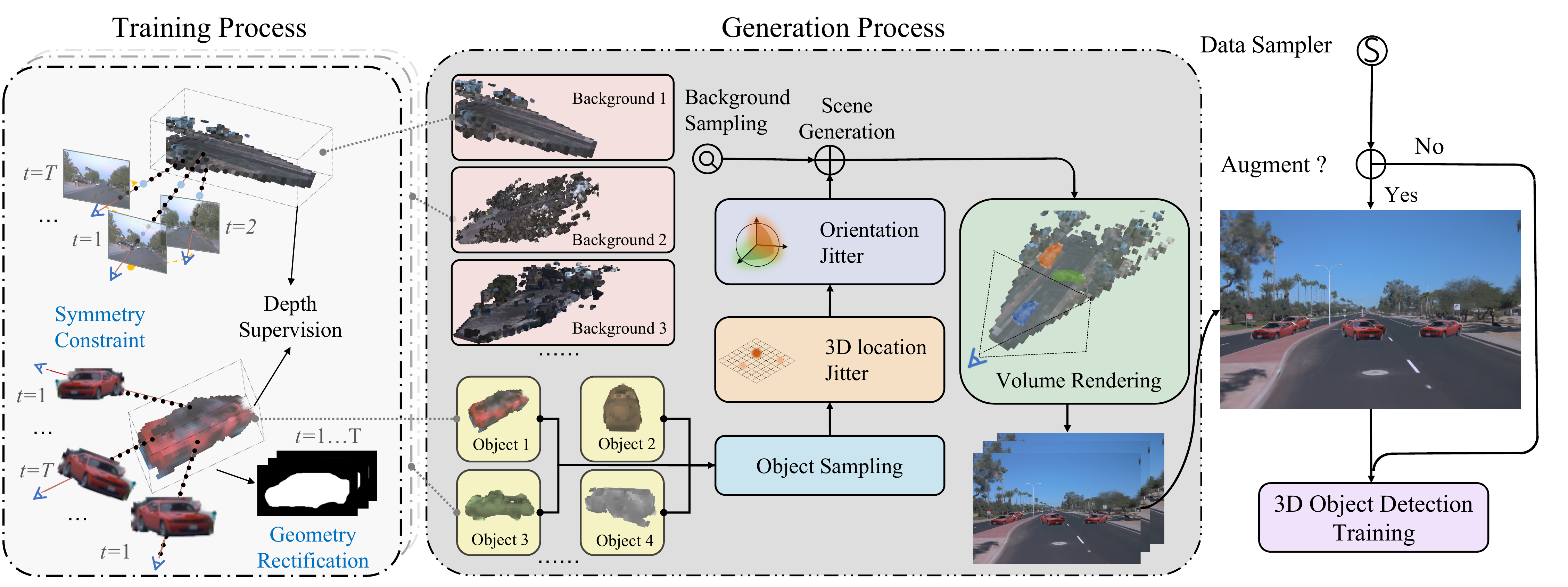}
  \caption{\textbf{Overview of Drive-3DAug} for 3D data augmentation. Driving scenes are decomposed into multiple backgrounds and objects. For each background and object, we use multi-frame views to reconstruct them separately by voxel-based NeRF \cite{sun2021direct}. To further improve the reconstruction quality, we introduce the symmetry constraint, geometry rectification, and depth supervision for the NeRF. We edit the scene in 3D space with the manipulation of trained 3D models, and the images are generated by rendering the composed new scenes for the following 3D perception tasks.}
  \label{fig:pipeline}
\end{figure*}


\section{Related Work}

\paragraph{Data Augmentation for 3D perception.} 
Data augmentation is a powerful technique to improve performances of perception algorithms~\cite{shorten2019survey, zoph2020learning}. For 3D perception, augmentation methods vary in data modalities. For point cloud data, flipping, rotation and translation of objects and backgrounds are common techniques~\cite{cheng2020improving, hahner2020quantifying, yan2018second, choi2021part}. For image data, 2D augmentation methods can be lifted to 3D with geometry constraints. \cite{lian2022exploring} improve random scale, crop and copy-paste from 2D to 3D with 2D--3D geometry relationship. For multi-modality data, \cite{zhang2020exploring, wang2021pointaugmenting} are proposed to keep a consistency between images and point clouds. 
Although these methods improve the performance of 3D perception tasks, the operations on the image plane are much less flexible than the 3D data augmentation for point clouds, \textit{e.g.}, unable to rotate vehicles. Besides, aggressive augmentation techniques on images always violate geometry constraints and lead to unnatural, flawed data. In comparison, the approach we proposed can generate images of diverse driving scenes with more degrees of freedom by object manipulation in the 3D space, which takes geometry constraints and occlusions into account at the same time.

\paragraph{NeRF for Scene Generation.}

NeRF \cite{mildenhall2020nerf} is a powerful tool for novel view synthesis, which represents a scene with a fully connected neural network and optimizes it with differentiable volume rendering. It has been recently applied to autonomous driving scenarios \cite{tancik2022block, li2022read, muller2022autorf, fu2022panoptic}. Block--NeRF~\cite{tancik2022block} reconstructs a whole city by merging multiple block-NeRFs with predicted visibility. 
Auto-RF~\cite{muller2022autorf} focuses on vehicle reconstruction in autonomous driving scenes. 
Considering the majority of NeRF methods are per-scene fitting, how to quickly reconstruct a scene is an important challenge because of the large scale of driving scenes. To overcome this, voxel-based NeRF~\cite{yu2022plenoxels, sun2021direct, mueller2022instant} and depth-supervised NeRF~\cite{deng2022depth}, which are adopted in our method, have been proposed to accelerate the training of NeRF. Moreover, some works begin to explore how to edit scenes with NeRF~\cite{OstMTKH21,KunduGYFPGTDF22,li2022read} for autonomous driving. 
However, these works are not extended to augmenting the data for 3D perception tasks. PNF~\cite{KunduGYFPGTDF22} does not consider driving data only has few views for most objects, which restricts the quality of novel view synthesis.
READ~\cite{li2022read} realizes autonomous driving scene editing with larger viewpoints of objects by point-based NeRF. However, this method relies on LiDAR data which limits its application scope. By contrast, the LiDAR data is not necessary for our method.



\section{Method}

Our goal is to create diverse driving scenes for improving 3D perception tasks, especially for camera-based 3D detection, by aid of 3D data augmentation. To this end, we propose 
Drive-3DAug. 

\subsection{Overview}~\label{overview}
As demonstrated in Figure \ref{fig:pipeline}, Drive-3DAug has two stages for implementing 3D data augmentation. 
\paragraph{Stage 1 - Training.} The first stage is to construct the 3D models from images for the following generation of new driving scenes. To achieve this, we decompose the scenes into backgrounds and foregrounds and build the 3D models of them by training the NeRF. Then we can edit the scenes in the 3D space. Considering the characteristics of driving data, we further develop several techniques to improve the training process. 
\paragraph{Stage 2 - Generation.} The second stage is to augment the training data through the 3D models. 
To create new driving scenes, we combine the models of foreground objects and background scenes in the 3D space with the manipulation of objects including the location and orientation jittering. 
In order to make the generated scenes close to the real-world driving scene, e.g., we cannot place vehicles on a tree, we design a strategy to identify the valid region of the background scene where we place the foreground object. Then we use volume rendering to generate new images of these scenes for the training of camera-based 3D perception tasks. 

\subsection{3D Model Training}~\label{reconstruct}
To edit driving scenes for 3D data augmentation, we need to construct a set of 3D models for backgrounds and foregrounds. To achieve this, we first utilize an off-the-shelf instance segmentation model to extract the objects from the backgrounds. Then we use the Intersection of Union (IoU) of the object masks and the projection of 3D boxes on images to match the extracted object and the 3D annotation.
After this, we use NeRF to reconstruct the 3D models from images. 
We reconstruct the 3D objects based on the extracted masks in consecutive frames by matching the object masks with IoU constraint, and we only consider the totally visible objects with intact masks for 3D reconstruction. 
To model the static backgrounds, the moving object masks in the images are filtered.

To efficiently reconstruct the backgrounds and objects, we use voxel-based NeRF \cite{yu2022plenoxels,sun2021direct,mueller2022instant} instead of MLP-based NeRF \cite{mildenhall2020nerf,park2021nerfies}. Voxel-based NeRF has a density voxel grid $\boldsymbol{V}_{\text{density}}$, and a feature voxel grid $\boldsymbol{V}_{\text{color}}$ with shallow MLPs to represent the geometry and appearance respectively. 
The NeRF model is trained by minimizing the loss between the rendered pixel color $\boldsymbol{C}(\boldsymbol{r})$ and observed pixel color $\hat{\boldsymbol{C}}(\boldsymbol{r})$ along the ray $\boldsymbol{r}$ given by
\begin{equation}
\mathcal{L}_{\text {Color }}=\sum_{\boldsymbol{r} \in \mathcal{R}(\mathbf{P})}\|\hat{\boldsymbol{C}}(\boldsymbol{r})-\boldsymbol{C}(\boldsymbol{r})\|_2^2,
\end{equation}
where $\mathcal{R}(\mathbf{P})$ is the set of rendered rays in a batch. 
To accelerate the model training, we introduce depth supervision to optimize the voxel field.
Then the NeRF model is trained by minimizing the loss
\begin{equation}
\mathcal{L} = \mathcal{L}_{\text {color}} + \mathcal{L}_{\text {Depth }},
\end{equation}
where $\mathcal{L}_{\text {Depth }}$ is the $L_1$ loss between the rendered depth and observed depth.
We can use LiDAR data or estimated depth by methods such as as SFM \cite{schonberger2016structure} or PACKNet \cite{packnet} to obtain the observed depth. We adopt the LiDAR data in this work as it is common in driving data.
For the convenience of scene editing, we reconstruct the static background of the scene in the world coordinate system and the object modeling the local object 3D box coordinate system.
Once we finish the training of the background and object NeRFs, they can serve as the digital driving assets for the repeated generation of novel scenes.



\paragraph{Analysis.}
Although the 3D models trained through this progress can be used for novel scene creation, they still struggle with several problems. First, because of the complexity of the driving scenes and the imperfect instance segmentation model, the extracted images of objects usually suffer from edge defects. 
NeRF can not model the object geometry well based on the noise mask as the background pixel may leak into the object model. 
In addition, the voxel grid representation through interpolation in voxel-based NeRF can further hinder the modeling of clear geometry near object boundaries.
Second, the range of available viewpoints for objects is very important for creating diverse driving scenes.
However, the objects in the driving data have limited views, indicating that we can only render objects within a small degree. This will restrict the diversity of the generated driving scenes. 
In terms of this, we refine the voxel-based NeRF model to promote the object models. 

\begin{figure}
  \centering
  \includegraphics[width=1.0\linewidth]{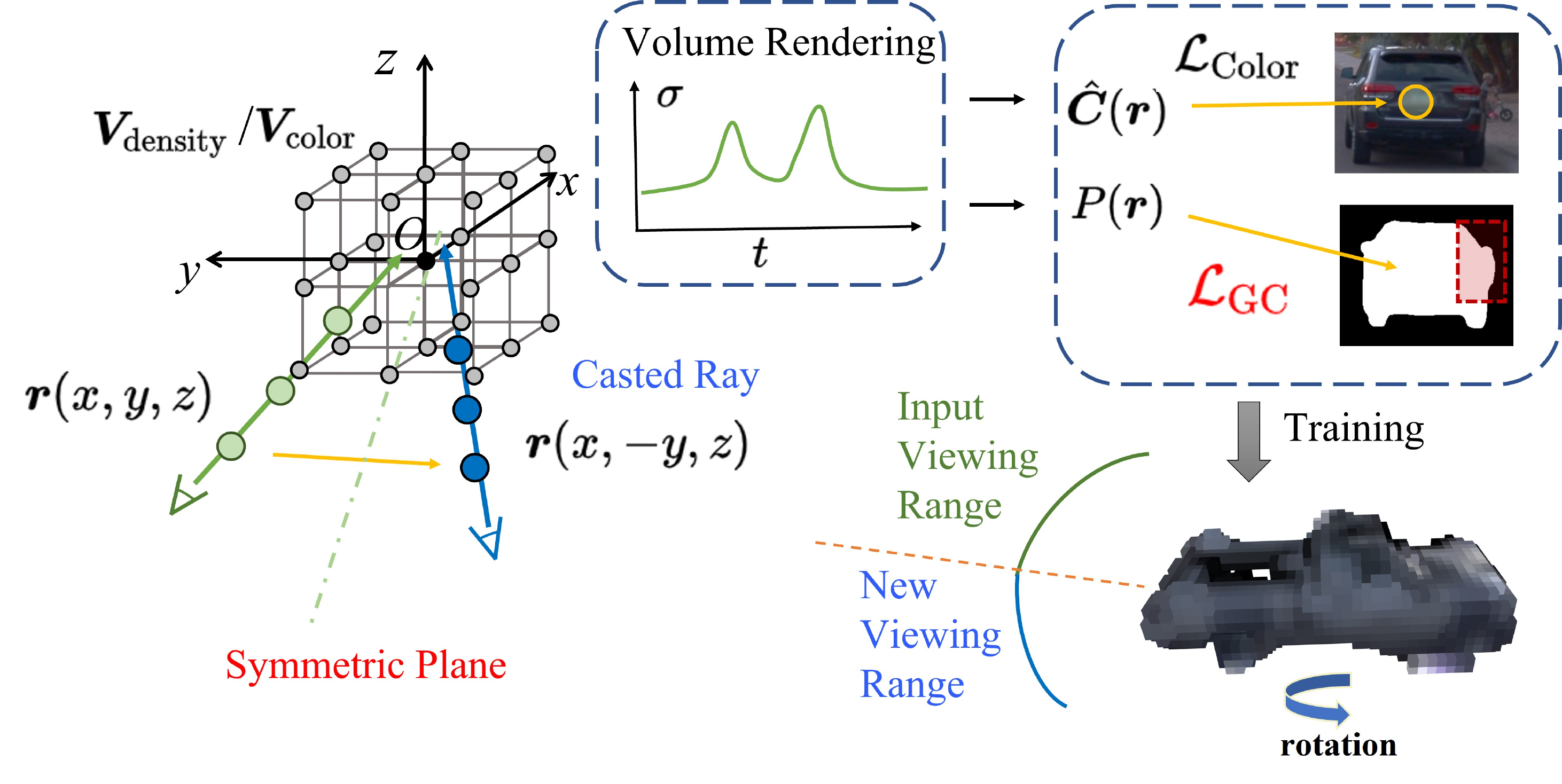}
  \caption{\textbf{Modified NeRF} including a geometric rectified loss and a symmetric-aware training strategy. They can alleviate the effects of imperfect object extraction and increase the range of viewpoints, respectively.}
  \label{fig:module}
\end{figure} 

\subsection{Improvements of Object Models}~\label{improve}
We design a geometric rectified loss and a symmetric-aware training strategy for the object models to ensure the quality and diversity of the generated novel scenes.

\paragraph{Geometric Rectified Loss.}
To avoid the effects of imperfect object extraction, we propose a geometric rectified loss to correct the geometry of the object model. As illustrated in Figure \ref{fig:module}, we add an auxiliary task for the training of the NeRF model by the classification of the pixel as the foreground or background pixel. Because the edge defects are different across consecutive frames, the temporal inconsistency can make the model remove the edge defects.
Specifically, the probability of a rendered pixel being an object pixel can be approximated by
\begin{equation}
P(\boldsymbol{r})= 1 - \exp \left(-\int_{t_a}^{t_b} \sigma  d s\right),
\end{equation}
where $t_a$ and $t_b$ denote the entrance and exit points of the ray-object intersection respectively, and $\sigma$ is the density of sampled point in the ray. Then we implement the geometric rectified loss as
\begin{equation}
\mathcal{L}_{\text {GC}}= -\sum_{\boldsymbol{r} \in \mathcal{R}(\mathbf{P})}log P(r) ,
\end{equation}
to decrease the voxel density near the mask edge, where $\mathcal{R}(\mathbf{P})$ is the set of rendered rays in a batch.
The final loss for training the object voxel field is defined by
\begin{equation}
\mathcal{L}_{\text{object}} = \mathcal{L}_{\text {Color }} + \mathcal{L}_{\text {Depth }} + \mathcal{L}_{\text {GC}}.
\end{equation}

\paragraph{Symmetric-aware Training Strategy.}
To enrich the viewpoints of objects, we design a symmetric-aware training strategy. Considering that objects, such as vehicles, are usually geometric symmetric in the driving scenes, we can create a symmetric object voxel field for them.
As depicted in Figure \ref{fig:module}, given a pixel $\boldsymbol{p}_r$ along the ray $\boldsymbol{r}(x, y, z)$, we can create a symmetric virtual camera that casts a ray $\boldsymbol{r}(x, -y, z)$ and targets the pixel $\boldsymbol{p}_l$, in which
$\boldsymbol{p}_r = \boldsymbol{p}_l$, supposing the symmetry of the object.
In this way, we increase the number of camera views for the object model training.
Then we can render novel views of the object rotated with a larger degree.

\subsection{Generation of Augmented Driving Scenes}\label{generation}

\begin{figure}
  \centering
    \includegraphics[width=0.98\linewidth]{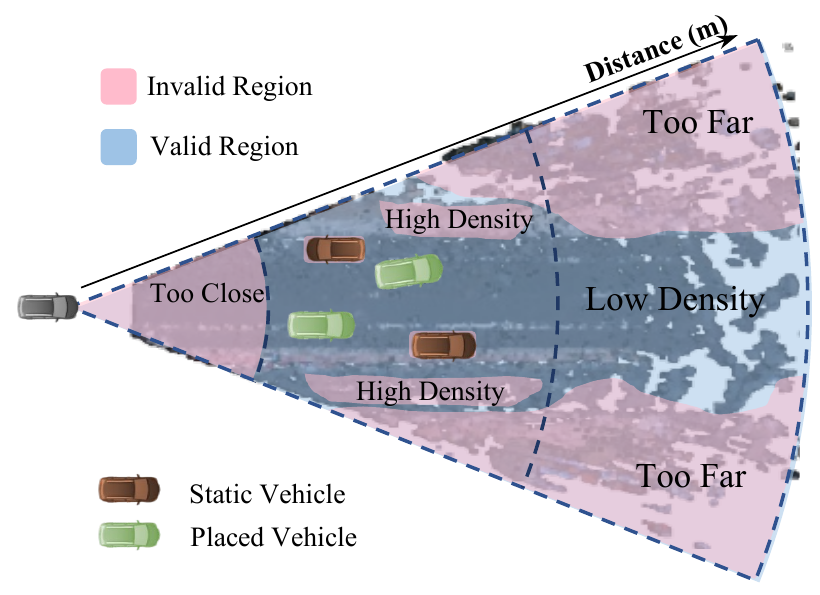}
  \caption{\textbf{Valid region} identification for the placement of vehicles, when composing a new driving scene.}
  \label{fig:validregion}
\end{figure}
As indicated in Figure \ref{fig:pipeline}, we generate a new driving scene through the combination of the object models and background models. We can place objects following the original location and orientation distributions of objects in data or change the distributions, such as making samples more balanced.
Besides, to make the placement of objects satisfy the physical law or other constraints in the real world, such as car on the road instead of the sky, we also identify the valid region of the background before the placement. Otherwise, the models of 3D perception tasks cannot learn the proper context information.  As shown in Figure \ref{fig:validregion}, we define the valid region on the bird's eye view based on the density field of the background model, which is inspired by the Lidar-Aug \cite{fang2021lidar}. To find the valid region, we first divide the 3D space into sets of pillars, and then the region can be split into valid and invalid states based on the density distribution of voxel in the corresponding pillar. 
Specifically, the valid region satisfies the low-density constraints: 
\begin{equation}
\max(Z_p) < \delta_1,  \, \text{and} \,\, mean(Z_p) < \delta_2,
\label{eq:valid_region}
\end{equation}
where $Z_p$ is the array denoting the density of points in the pillar, and $\delta_1$ and $\delta_2$ are hyper-parameters. This represents no large object is in this region.
Furthermore, we can filter the low-density region behind the high-density region such as the wall. 
In this way, we can put the object in an appropriate position.  
To avoid the collision, we calculate 3D IoUs between the placed objects and existing objects to ensure consistency between the foreground objects and the backgrounds. 
Then we jointly render the merged 3D model to generate images of the new scenes. 
The generated images can be directly applied for the downstream 3D detection task to improve the performance of detectors.

\section{Experiments}

We validate the proposed Drive-3DAug method on the monocular 3D detection task, which is one of the most important 3D perception tasks in autonomous driving.

\subsection{Datasets and Metrics}\label{subsec:dataset}

\paragraph{Waymo Dataset.}
The Waymo Open Dataset \cite{sun2020scalability} is a large-scale dataset for autonomous driving that contains 798 scenes in the training dataset and 202 scenes in the validation dataset. The image resolution for the front camera is $1920 \times 1280$. Waymo uses the LET-AP \cite{hung2022let}, the average precision with longitudinal error tolerance, to evaluate detection models. Besides, Waymo also adopts the LET-APL and LET-APH metrics, which are the longitudinal affinity weighted LET-AP and the heading accuracy weighted LET-AP, respectively.
    

\paragraph{nuScenes Dataset.}

The nuScenes \cite{caesar2020nuscenes} is a widely used benchmark for 3D object detection.
It contains 700 training scenes and 150 validation scenes. The resolution of each image is 1600 $\times$ 900.  As for the metrics, nuScenes computes mAP using the center distance on the ground plane to match the predicted boxe and the ground truth. It also contains different types of true positive metrics (TP metrics). We use ATE, ASE and AOE in this paper, for measuring the errors of translation, scale and orientation, respectively.

\setlength{\tabcolsep}{10pt}
\begin{table*}[t]
\small
\begin{center}
\begin{tabular}{@{}l|ccc|cccc@{}}
  \toprule
  \multirow{2}{*}{\textbf{Method}} 
  & \textbf{LET-AP}  & \textbf{LET-APH}  & \textbf{LET-APL}  & \makecell[c]{\textbf{LET-APL}\\ $[50\mathrm{m},+\infty)$} & \makecell[c]{\textbf{LET-APH}\\ $\mathbin{\sim}45^{\circ}$} & \makecell[c]{\textbf{LET-APH}\\ $\mathbin{\sim}90^{\circ}$} & \makecell[c]{\textbf{LET-APH}\\ $\mathbin{\sim}135^{\circ}$} \\ 
  \midrule
  FCOS3D \cite{wang2021fcos3d}            & 0.585 & 0.573 & 0.393 & 0.278 & 0.293 & 0.285 & 0.223 \\
  + Copy-paste \cite{lian2022exploring} & 0.594 & 0.581 & 0.401 & 0.290 & 0.295 & 0.283 & \textbf{0.234} \\
  + Drive-3DAug w/o RT         & 0.595 & 0.583 & 0.404 & 0.287 & 0.293 & 0.295 & 0.226 \\
  + Drive-3DAug w/ RT         & \textbf{0.602} & \textbf{0.590} & \textbf{0.410} & \textbf{0.298} & \textbf{0.315} & \textbf{0.299} & \textbf{0.234} \\
  \midrule
  SMOKE \cite{liu2020smoke}            & 0.586 & 0.579 & 0.417 & 0.304 & 0.312 & 0.291 & 0.264 \\
  + Copy-paste \cite{lian2022exploring} & 0.594 & 0.587 & 0.425 & 0.320 & 0.303 & 0.301 & 0.239 \\
  + Drive-3DAug w/o RT & 0.592 & 0.584 & 0.421 & 0.318 & 0.314 & 0.298 & \textbf{0.266} \\
  + Drive-3DAug w/ RT   & \textbf{0.598} & \textbf{0.590} & \textbf{0.426} & \textbf{0.322} & \textbf{0.333} & \textbf{0.303} & 0.255 \\
\bottomrule
\end{tabular}
\end{center}
\caption{\textbf{Monocular 3D detection results} on the Waymo validation set. The left part shows the main metrics. LET-AP represents longitudinal error tolerant 3D average precision \cite{hung2022let}. LET-APH and LET-APL represent LET-AP penalized by heading errors and longitudinal localization errors, respectively. The right part of the table shows metrics under specific hard cases. The $[50\mathrm{m},+\infty)$ metric only calculates AP over objects with distance to ego-car greater than $50\mathrm{m}$. The $\mathbin{\sim} h ^{\circ}$ metric only consider the objects with $\vert \mathrm{heading} \vert$ near. It is observed that our \textit{Drive-3DAug w/ RT} performs consistently better than any other setting.}
\label{tab:main_results}
\end{table*}

\subsection{Implementation Details}\label{subsec:implementation}
\paragraph{Building Digital Driving Asset.} 
We use the SOLO v2~\cite{wang2020solo} trained on COCO as the instance segmentation model for scene decomposition. 
We consider \textit{vehicle} (Waymo) or \textit{car} (nuScenes) as the foreground objects, since they are the most important components in driving scenes 
For 3D model reconstruction, we use the same model configuration as the DVGO~\cite{sun2021direct} with our proposed techniques. We use 30-40 consecutive frames spanning an area of about 100-200 meters as one background and train each background model with 40,000 iterations. For the background voxel grid, we set the resolution as $330^3$ with a voxel size of 0.25-0.3m. The object models are trained with 20-60 consecutive frames, and we set the voxel size as 0.25m consistent with the background voxel size. The grid point number is about 1,000.
Considering the construction cost and limitation of NeRF model for extreme illumination conditions, we select a subset of 100 sunny scenarios for each dataset. We use them for the data augmentation.

\begin{figure}[t]
    \centering
    \includegraphics[width=0.99\linewidth]{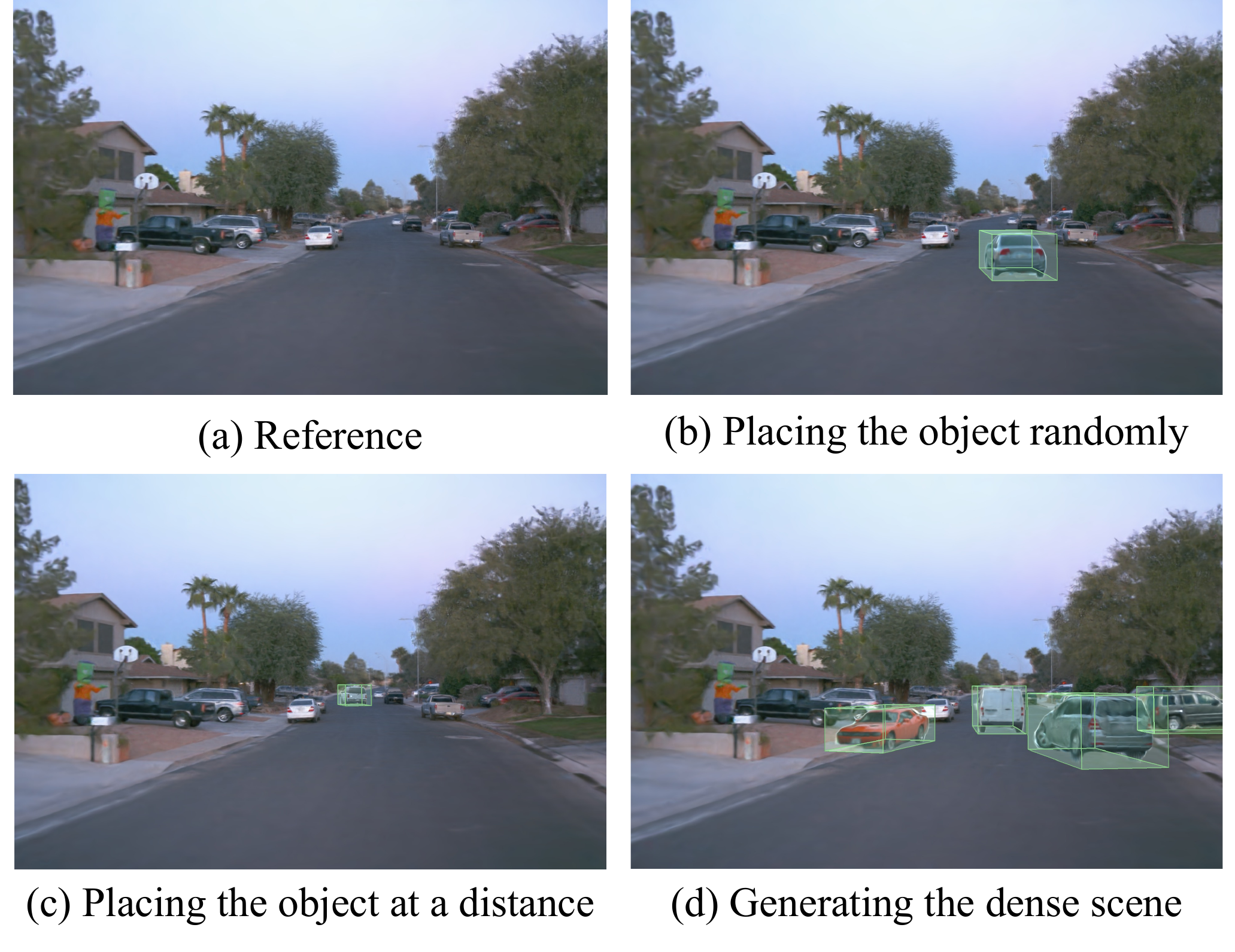}
    \caption{\textbf{Scene generation} with different placement strategies of the \textit{vehicle}.}
    \label{fig:gen_strategy}
\end{figure}

\paragraph{Applying Data Augmentation.}
Our method generates new data through rendering the recomposed scenes of the randomly selected 3D models. We manipulate the object in 3D space by 3D location jittering and orientation jittering and place it on the valid region of arbitrary backgrounds. 
The location jittering is defined by the maximum translation $(T_x, T_y)$ along the x-direction and the y-direction, and the orientation jittering is given by maximum rotation angle $T_{\theta}$. We consider two data augmentation strategies to valid the effectiveness of adding rotation and translation for 3D perception. The first is $\textbf{\textit{Drive-3DAug w/o RT}}$, in which we set translation $T_x=0$, $T_y=0$, and rotation $T_{\theta}=0$ and 1-2 new objects are arbitrarily pasted into the background scene on average. The second is $\textbf{\textit{Drive-3DAug w/ RT}}$ with $T_x=20$m, $T_y=5$m, and $T_{\theta}=30^{\circ}$. 
As for determining the valid region in the background scene, we set the pillar $Z_p$ resolution as 2m$\times$2m, $\delta_1=30$, and $\delta_2=15$ in Eq.~\ref{eq:valid_region}. 
We generate 12 new images for every background model. This progress is offline data augmentation and these images are repeatedly used by different detectors.



\begin{figure}[t]
    \begin{center}
    \includegraphics[width=0.989\linewidth]{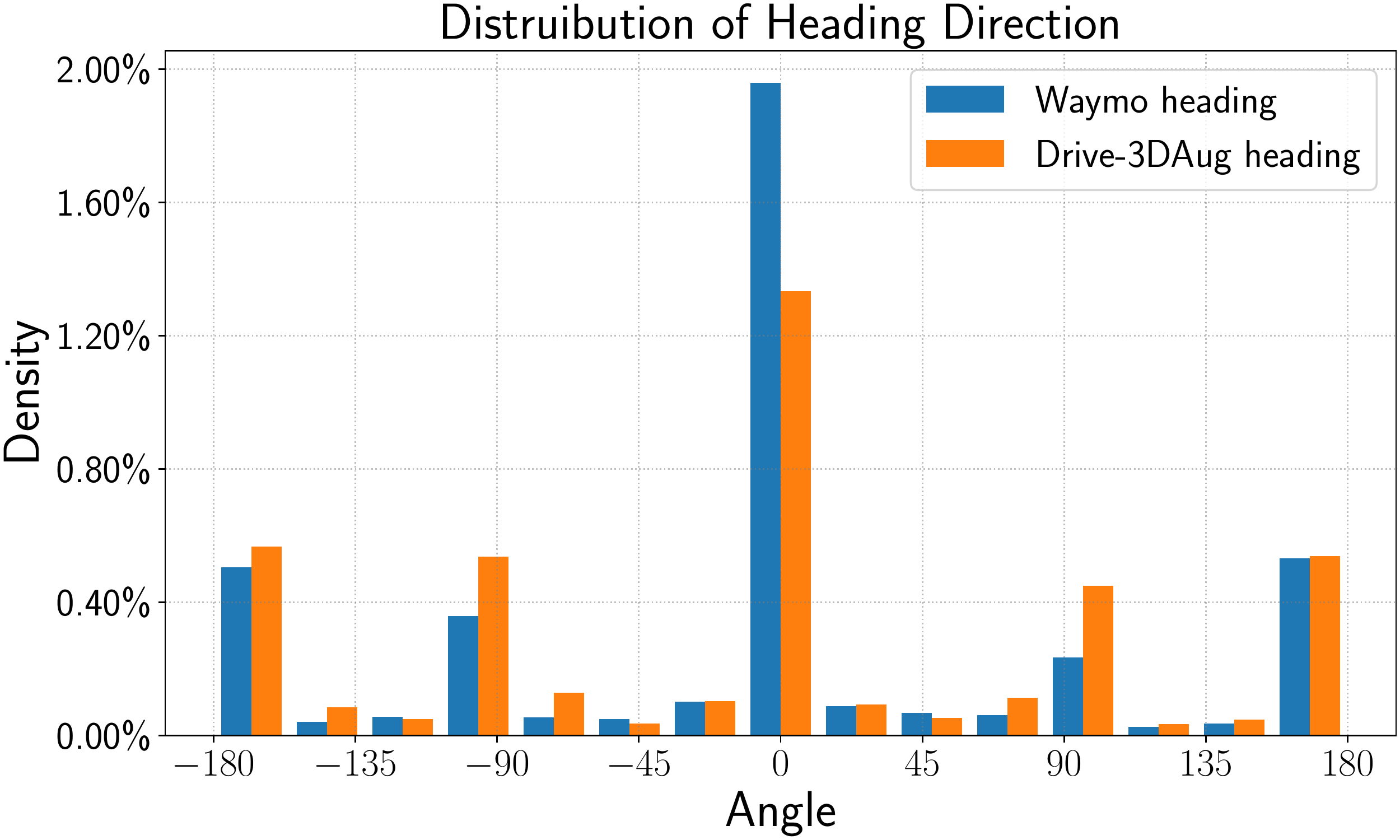}
    \caption{\textbf{Distribution of heading direction} for \textit{vehicle} on the Waymo training set and the distribution augmented by Drive-3DAug.}
    \label{fig:heading_distribution}
    \end{center}
\end{figure}

\paragraph{Training Detectors.}
Two typical camera-based monocular 3D object detectors, FCOS3D\cite{wang2021fcos3d} and SMOKE\cite{liu2020smoke} are utilized to investigate the performance of our proposed method as they are two of the most popular and commonly used mono3D detectors. We maintain the same hyperparameters of detectors for all experiments in our study, introduced in the supplementary. We use all the scenes in the training set to train the detectors and evaluate the detectors on the entire validation set. However, we sample training data every 3 frames on Waymo during training due to limited computational resources.
Besides, although our method can be applied to any view of cameras, we only take into account images taken by the front camera on Waymo and nuScenes in this work because of the computational resource.
For the usage of the generated images, we randomly replace the image in a batch with the augmented data if it belongs to the scenes in the digital driving asset.

\subsection{Main Results}\label{subsec:results}
\paragraph{Novel Scene Generation.}
Drive-3DAug can generate diverse scenes with the ability to manipulate objects in the 3D space. 
As indicated in Figure~\ref{fig:gen_strategy}, our method can place the car at any distance or generate a very dense scene to augment the data. Moreover, previous image data augmentation methods are likely to put the car on the sky because they do not define the valid region. By contrast, our placement obeys the real world situation.
Besides the placement, our method can also alleviate the imbalance problem of vehicle orientation with orientation jittering to objects. Figure~\ref{fig:heading_distribution} shows the extremely imbalanced distribution of vehicle heading direction on the Waymo dataset, where most heading directions are at $0^{\circ}$ or $180^{\circ}$. In comparison, the distribution of vehicle heading distribution is more balanced than the original distribution, although the augmented distribution is still imbalanced. This is because of the randomly jitter for the orientation of objects.

\setlength{\tabcolsep}{9pt}
\begin{table}[t]
  \small
  \begin{center}
  \begin{tabular}{@{}lccccc@{}}
    \toprule
    Method & AP  & ATE  & ASE  & AOE  \\
    \midrule
    FCOS3D\cite{wang2021fcos3d}   & 0.319 & 0.739 & 0.160 & 0.096 \\
    + Copy-paste \cite{lian2022exploring}             & 0.324 & 0.721 & \textbf{0.156} & 0.123 \\
    + Drive-3DAug w/ RT              & \textbf{0.333} & \textbf{0.705} & 0.158 & \textbf{0.092} \\
    \bottomrule
  \end{tabular}
  \end{center}
  \caption{\textbf{Monocular 3D detection results} on the nuScenes validation set. ATE, ASE, and AOE are used for measuring translation, scale, and orientation errors respectively.
  }
  \label{tab:nuscenes}
\end{table}

\begin{figure}[t]
    \centering
    \includegraphics[width=0.98\linewidth]{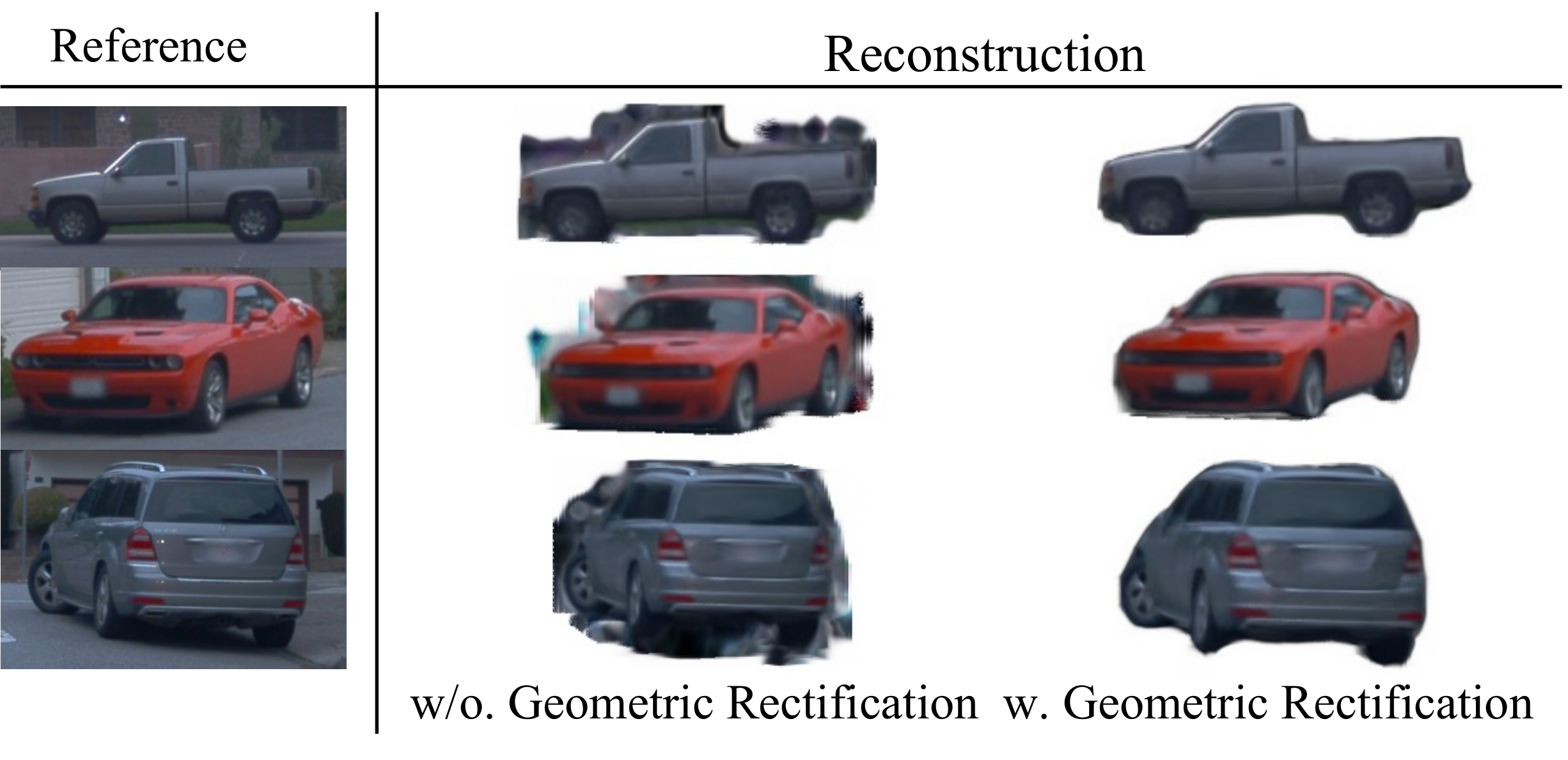}
    \caption{\textbf{Visualization} of the ablative study for the geometric rectified loss. After applying such a loss, the reconstructed object models have much fewer edge defects.}
    \label{fig:ablation_geo}
\end{figure}

\paragraph{3D Object Dectection.}
Table~\ref{tab:main_results} illustrates the main results on the Waymo validation set. We compare our Drive-3DAug with the geometry-consistent copy-paste method \cite{zhang2020exploring}, which copies and pastes objects across different scenes with depth constraints and camera transformation of 3D annotations. Besides, we also apply the valid region strategy for this method with the help of LiDAR data. It also uses the selected 100 scenes for augmentation.
Since our data augmentation method is applied only to vehicles, we use the LET-AP on \textit{vehicle} to evaluate the effectiveness of our approach. Compared to \textit{baseline}, our \textit{NeRF w/ RT} approach significantly improves LET-AP by 1.5\%, LET-APH by 1.4\%, and LET-APL by 1.5\% on average across both FCOS3D and SMOKE. The consistent improvement between these detectors indicates that our approach is robust to different backbone and head architectures. It also outperforms \textit{Copy-paste} by 0.6\%, 0.6\%, and 0.5\% on LET-AP, LET-APH, and LET-APL, respectively.  Furthermore, we also compare the performance of different methods under far distances and different orientation ranges of vehicles. The right of Table~\ref{tab:main_results} shows that the detector trained with \textit{Drive-3DAug w/ RT} outperforms other methods under all the scenarios, especially for the objects with $\mathbin{\sim}45^{\circ}$ orientation, which is the fewest in the original data. This means the ability to add larger translation and rotation to placed objects is essential for the data augmentation techniques in 3D perception.

Table~\ref{tab:nuscenes} reports the experimental results on the nuScenes validation set. We can see \textit{Drive-3DAug w/ RT} also improves the performance of class \emph{car} on the nuScenes dataset. 
Additionally, it is worth noting that Drive-3DAug obtains a $1.6\%$ improvement on ATE, a $3.1\%$ improvement on AOE for the FCOS3D detector compared with the geometry-consistent copy-paste method. The result on ASE is comparable for these two methods. This is because we do not apply the scale transformation. This further proves that our method is capable of resolving the issues of previous methods with only 2D pixel movements that lack the ability to generate realistic rotated or translated objects.


\setlength{\tabcolsep}{8.5pt}
\begin{table}[t]
  \small
  \begin{center}
  \begin{tabular}{@{}ccccc@{}}
    \toprule
     GRL  & SAT & LET-AP &  LET-APH & LET-APL \\
    \midrule
    -        &  -           &  0.590 & 0.578 & 0.403 \\
     \checkmark & - &  0.596 & 0.583 & 0.407 \\
    \checkmark & \checkmark &  \textbf{0.602} & \textbf{0.590} & \textbf{0.410} \\
    \bottomrule
  \end{tabular}
  \end{center}
  \caption{\textbf{Ablation Study} of Drive-3DAug w R/T for FCOS3D on Waymo validation set. GRL means the geometric rectified loss and SAT means symmetric-aware training.}
  \label{tab:ablation_nerf_model}
\end{table}

\begin{figure}[t]
    \centering
    \includegraphics[width=0.98\linewidth]{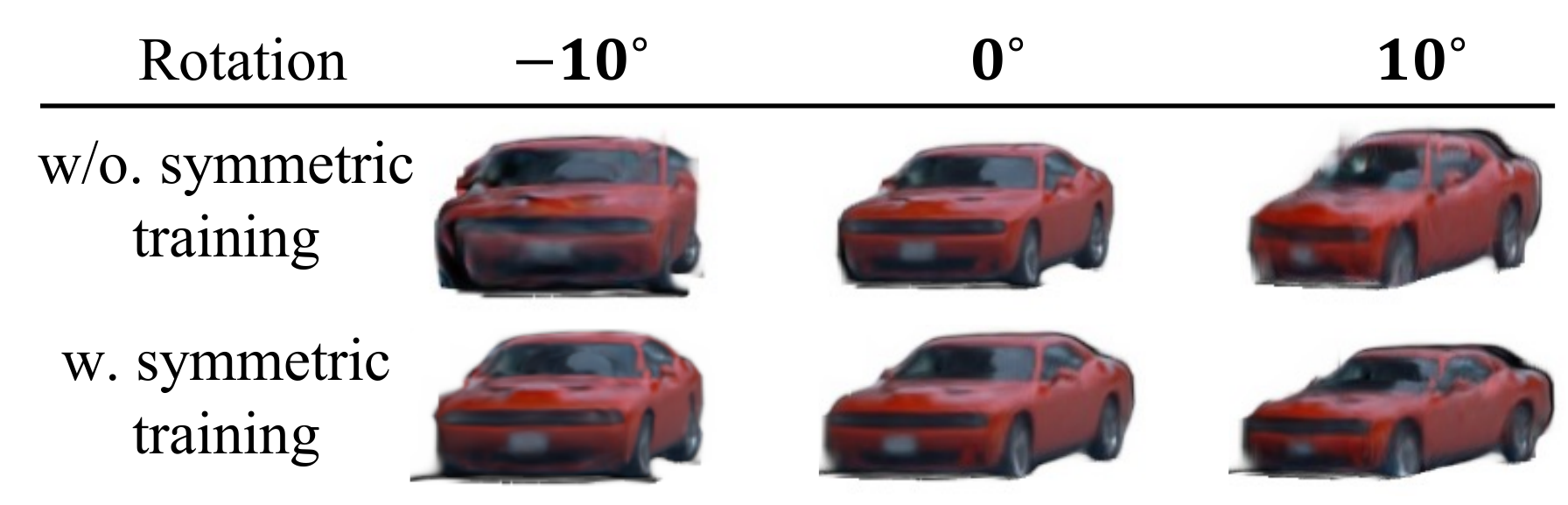}
    \caption{\textbf{Visualization} of the ablative study for the symmetric-aware training strategy. It can make the model generate more realistic images with larger degrees of rotation.}
    \label{fig:ablation_sym}
\end{figure}



\subsection{Ablative Studies}\label{subsec:analysis}


Table~\ref{tab:ablation_nerf_model} reports the results of our ablation experiments with the FCOS3D detector. The baseline is using DVGO \cite{sun2021direct} with depth supervision for data augmentation. The results indicate that even with vanilla voxel-based NeRF without any improvement, our proposed Drive-3DAug can improve the performance of the detector by 0.5\% on LET-AP.  After adding the geometric rectified loss and the symmetric-aware training strategy, the LET-AP is further increased by 0.6\% and 0.6\% respectively. In addition, we provide some visualizations to demonstrate the effectiveness of each component. 
Figure~\ref{fig:ablation_geo} shows that the geometric rectified loss can significantly suppress the edge defects and improve the reconstruction quality.
Figure~\ref{fig:ablation_sym} shows the results of the symmetric-aware training strategy for the novel view synthesis. We can see that adding this training strategy makes the model have larger viewpoints.

\begin{figure}[t]
  \centering
\includegraphics[width=0.99\linewidth]{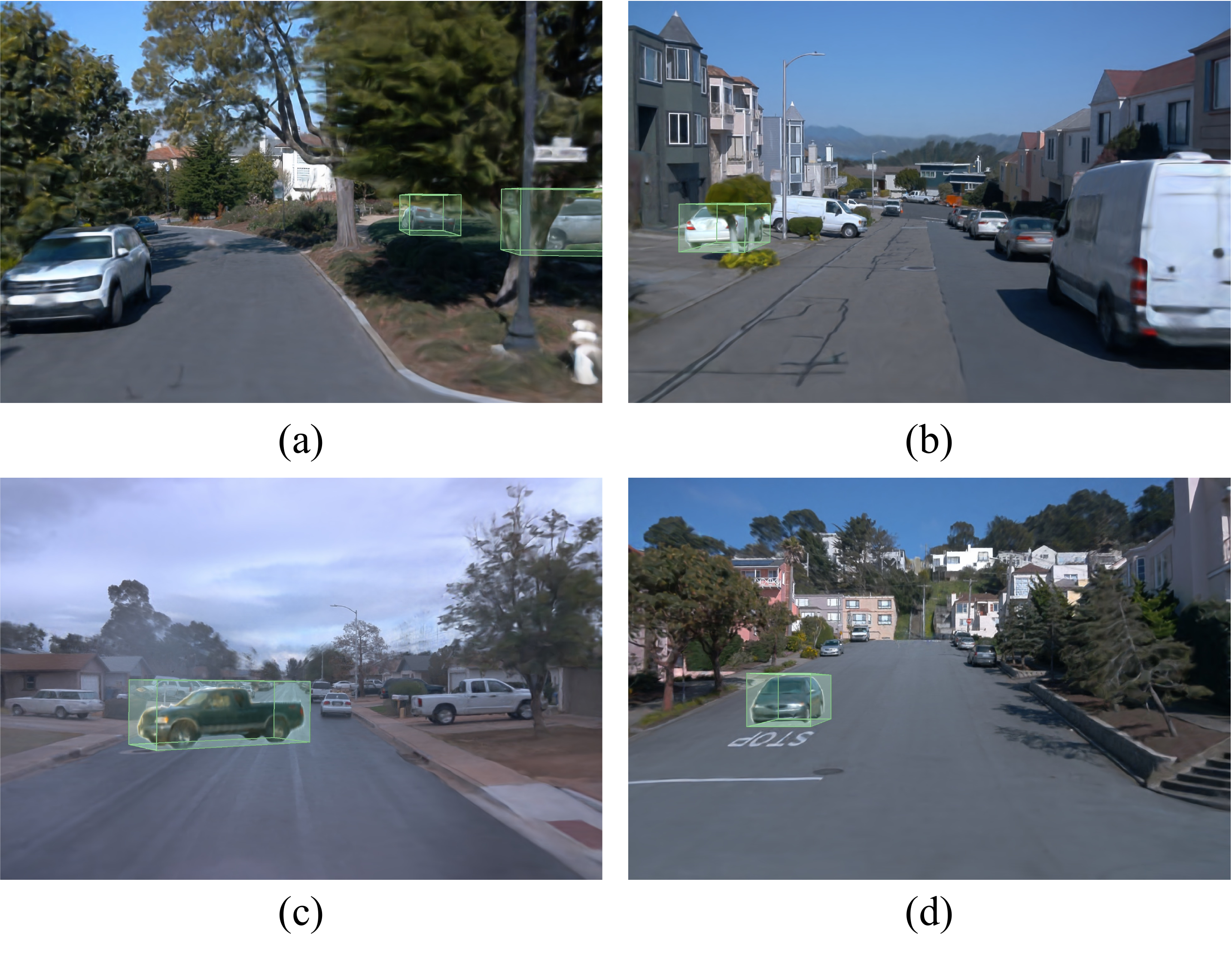}
\caption{\textbf{Corner Cases} generated by the Drive-3DAug, where (a), (b) illustrate occlusion situations, (c) demonstrates strange vehicle heading direction, and (d) shows a vehicle appended on slope road. Cars with 3D bounding boxes are rendered by NeRF.}
\label{fig:corner_case}
\end{figure}

\vspace{-2mm}
\paragraph{Reconstruction Cost.}
Table~\ref{tab:scece_recon_cost} depicts the comparison of reconstruction speed on a NVIDIA V100 GPU between previous methods and our method. Unlike the MLP-based NeRF~\cite{mildenhall2020nerf} which needs more than 20 hours of training for one background, it takes about 0.5h for the voxel-based NeRF with depth supervision. For the object model, the reconstruction time is within minutes. Considering the model size of our NeRF is rather small, we can run multiple reconstructions in parallel. 
Moreover, once these models are trained, they could be recycled for different detectors as a digital driving assets.

\setlength{\tabcolsep}{10pt}
\begin{table}[t]\small
  \begin{center}    
{
  \begin{tabular}{@{}lccc@{}}
    \toprule
    Method & Voxel Field.  &  Depth Sup.&    Cost \\
    \midrule
    NeRF~\cite{mildenhall2020nerf}  & -            & - & $> 20$h \\
    DVGO~\cite{sun2021direct}       & $\checkmark$ & - & $0.7$h \\
    Ours                            & $\checkmark$ & $\checkmark$ & \textbf{$0.5$h} \\
    \bottomrule
  \end{tabular}
  }
  \end{center}
  \caption{\textbf{Reconstruction cost} of different methods for one background, assessed on a NVIDIA V100 GPU with 16GB memory.
  }
  \label{tab:scece_recon_cost}
\end{table}

\vspace{-2mm}
\paragraph{Corner Case Generation.}
Drive-3DAug is able to generate many photographic data for various corner cases~\cite{cornercase2021} without much effort for autonomous driving systems. As described in Figure~\ref{fig:corner_case}, we use our method to simulate several corner cases including \emph{the car occluded by the environment}, \emph{the car appearing on the road with strange positions and headings}, and \emph{the car on the slope}, which are hard to collect in the real world. This shows that 3D data augmentation can help alleviate issues of autonomous driving caused by plenty of corner cases.



\setlength{\tabcolsep}{12pt}
\begin{table}[t]
\small
\begin{center}
\begin{tabular}{lcc}
\toprule
\multicolumn{1}{l}{\multirow{2}{*}{Method}} & \multicolumn{2}{c}{LET-AP} \\ \cmidrule{2-3}
  & Pedestrian & Cyclist \\ \midrule
FCOS3D \cite{wang2021fcos3d}                                        & 0.339         & 0.231             \\ 
+ Drive-3DAug w RT                           & \textbf{0.344}     & \textbf{0.234}        \\ \bottomrule
\end{tabular}
\end{center}
\caption{\textbf{Monocular 3D detection results} of pedestrian and cyclist on Waymo validation set. LET-AP is the longitudinal error tolerant 3D average precision.}
\vspace{-2mm}
\label{tab:pb}
\end{table}


\vspace{-3mm}
\paragraph{Pedestrian and Cyclist Augmentation.}

Table \ref{tab:pb} shows the results for applying Drive-3DAug to the \textit{pedestrian} and \textit{cyclist} on Waymo. These objects are difficult for reconstruction, especially in driving scenes. Because they are not rigid objects and may change their pose dramatically in consecutive frames. However, our method can still improve the detection results for these two classes by 0.5\% and 0.3\% on LET-AP, respectively.


\section{Conclusion}

In this paper, we propose Drive-3DAug, the first 3D data augmentation technique for camera-based 3D perception task in autonomous driving. We represent the scene with background and foreground 3D models by NeRF and randomly combine them to generate new driving scenes. 
The quality and diversity issues of generated scenes are addressed by our novel geometric rectified loss and symmetry-aware training strategies.  
We demonstrate the effectiveness of our method on multiple datasets and detectors.  Furthermore, these 3D models can be regarded as the digital driving asset and benefit the community of this area.

\vspace{-2mm}
\paragraph{Discussion.}
Driving scenes are extremely complicated, including lots of object categories under different illumination and weather conditions. Currently, our method only augments limited classes of objects under good illumination conditions. It is worth to include more situations as the digital driving asset for the future work. 

{\small
\bibliographystyle{ieee_fullname}
\bibliography{egbib}
}

\clearpage
\appendix

\section{Preliminaries of NeRF}\label{sec:priliminary}
The 3D scene can be represented with the NeRF model, and the neural network is utilized to map a point position $\boldsymbol{x} \in \mathbb{R}^3$  and a view direction $\boldsymbol{d} \in \mathbb{R}^3$ to the corresponding color $\boldsymbol{c} \in \mathbb{R}^3$ and volume density $\sigma$ \cite{mildenhall2020nerf}.  We apply the voxel grid to represent the scene considering the low computational cost of voxel-based NeRF \cite{sun2021direct}. The density voxel grid $\boldsymbol{V}_{\text{density}}$  and  feature voxel grid $\boldsymbol{V}_{\text{color}}$  with a shallow MLP  are adopted to represent the scene geometry and appearance, respectively. Given input queries $\boldsymbol{x} $ and $\boldsymbol{d} $,  the outputs are obtained with the interpolation
\begin{equation}
\begin{aligned}
 \sigma &= \text{inter}(\boldsymbol{x}, \boldsymbol{V}_{\text{density}})  \\
 \boldsymbol{c} &= \text{MLP}_{\theta}(\text{inter}(\boldsymbol{x}, \boldsymbol{V}_{\text{color}}), \boldsymbol{x} , \boldsymbol{d}) 
  \label{eq:important}
  \end{aligned}
\end{equation}
To render the image, the pixel color $\boldsymbol{C}(\boldsymbol{r})$ along the camera ray $\boldsymbol{r}(t) = \boldsymbol{r_0}+ t \boldsymbol{d}$ is approximated by the volume rendering
\begin{equation}
\boldsymbol{C}(\mathbf{r})=\int_{t_1}^{t_2} T(t) \sigma(\boldsymbol{r}(t)) \mathbf{c}(\boldsymbol{r}(t), \mathbf{d}) d t,
\end{equation}
where  $t_1$ and $t_2$ are near and far bounds for sampling points, $\boldsymbol{r_0}$ is the camera origin, and $T(t)$ is accumulated transmittance along the ray from $t_1$ to $t$ defined by
\begin{equation}
T(t)=\exp \left(-\int_{t_1}^t \sigma(\boldsymbol{r}(s)) d s\right).
\label{eq:transmittance}
\end{equation}
The NeRF model is trained by minimizing the  loss between the rendered pixel color $\boldsymbol{C}(\boldsymbol{r})$ and observed pixel color $\hat{\boldsymbol{C}}(\boldsymbol{r})$given by
\begin{equation}
\mathcal{L}_{\text {Color }}=\sum_{\boldsymbol{r} \in \mathcal{R}(\mathbf{P})}\|\hat{\boldsymbol{C}}(\boldsymbol{r})-\boldsymbol{C}(\boldsymbol{r})\|_2^2,
\end{equation}
where $\mathcal{R}(\mathbf{P})$ is the set of rendered rays in a batch.

\section{Implementation Detail of Detectors}
For FCOS3D, we utilize a ResNet-101-DCN\cite{he2016deep, dai2017deformable} as the backbone. The model is trained for 24 epochs using the SGD optimizer with an initial learning rate of 1e-4 and a momentum of 0.9. We set the weight decay to 1e-5, and the max norm of gradient clipping to 35. We also adopt a step decay learning rate scheduler with a 0.1$\times$ decrease at epoch 20 and 23, along with 1000 iterations of linear warm-up. For SMOKE, we employ a DLA-34\cite{yu2018deep} as the backbone. We use the Adam optimizer with an initial learning rate 1e-4, and the remaining settings are the same as FCOS3D. For both detectors, their backbones are initialized with ImageNet pre-trained weights. The batch size for training is set to 16.

\setlength{\tabcolsep}{8.5pt}
\begin{table}[t]
  \small
  \begin{center}
  \begin{tabular}{@{}ccccc@{}}
    \toprule
     3DAug  & DS & LET-AP &  LET-APH & LET-APL \\
    \midrule
      -     &  -           &  0.585 & 0.573 & 0.393 \\
     \checkmark  & - & 0.584 & 0.572   & 0.394 \\
    \checkmark & \checkmark &  \textbf{0.590} & \textbf{0.578} & \textbf{0.403} \\
    \bottomrule
  \end{tabular}
  \end{center}
  \caption{\textbf{Ablation Study} of Drive-3DAug for FCOS3D on Waymo validation set. 3DAug means we use DVGO \cite{sun2021direct} for data augmentation. DS means depth supervision.}
  \label{tab:ablation_depth}
\end{table}

\begin{figure}[t]
    \begin{center}
        \includegraphics[width=0.99\linewidth]{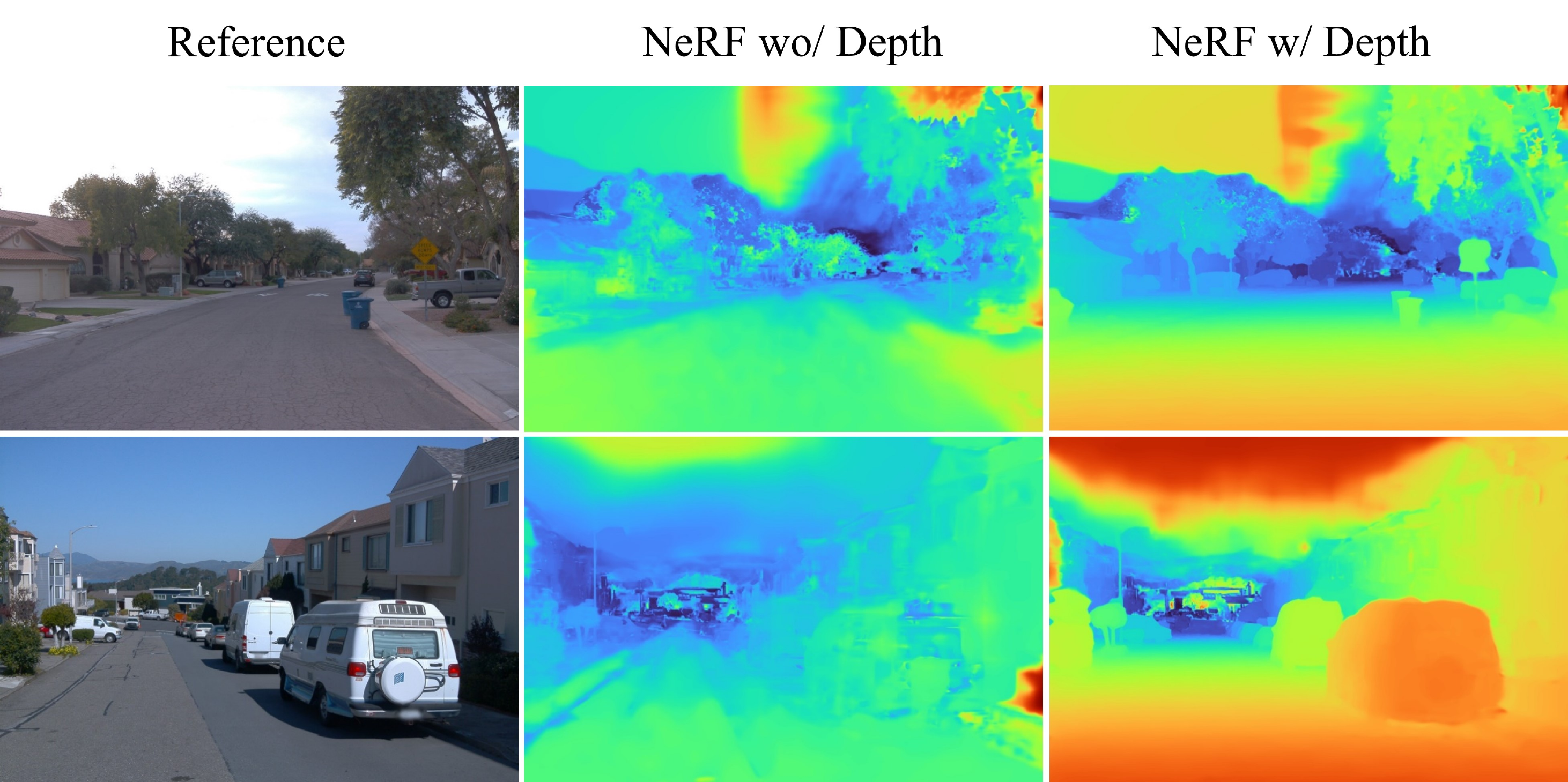}
    \end{center}
    \caption{\textbf{Visualization of rendered depth map.}  The model with depth supervision depicts better performance.}
    \label{fig:depth}
\end{figure}

\section{Ablation Study on Depth Supervision.}
We qualitatively and quantitatively investigate the effect of depth supervision on background model training and 3D augmentation.
Table~\ref{tab:ablation_depth} shows that the 3D augmentation based on background model trained with depth supervision has better performance, with LET-AP (0.590 vs 0.585).
Figure~\ref{fig:depth} shows that NeRF can reconstruct the background with high quality given depth supervision, and the 3D background model quality can be decreased without depth information.
Thus, LET-AP, LET-APH and LET-APL on car have a slight decrease with 0.001 for 3D augmentation using background model without supervision.

\section{Visualization of Drive-3DAug}
\begin{figure*}[t]
    \begin{center}
\includegraphics[width=0.9\linewidth]{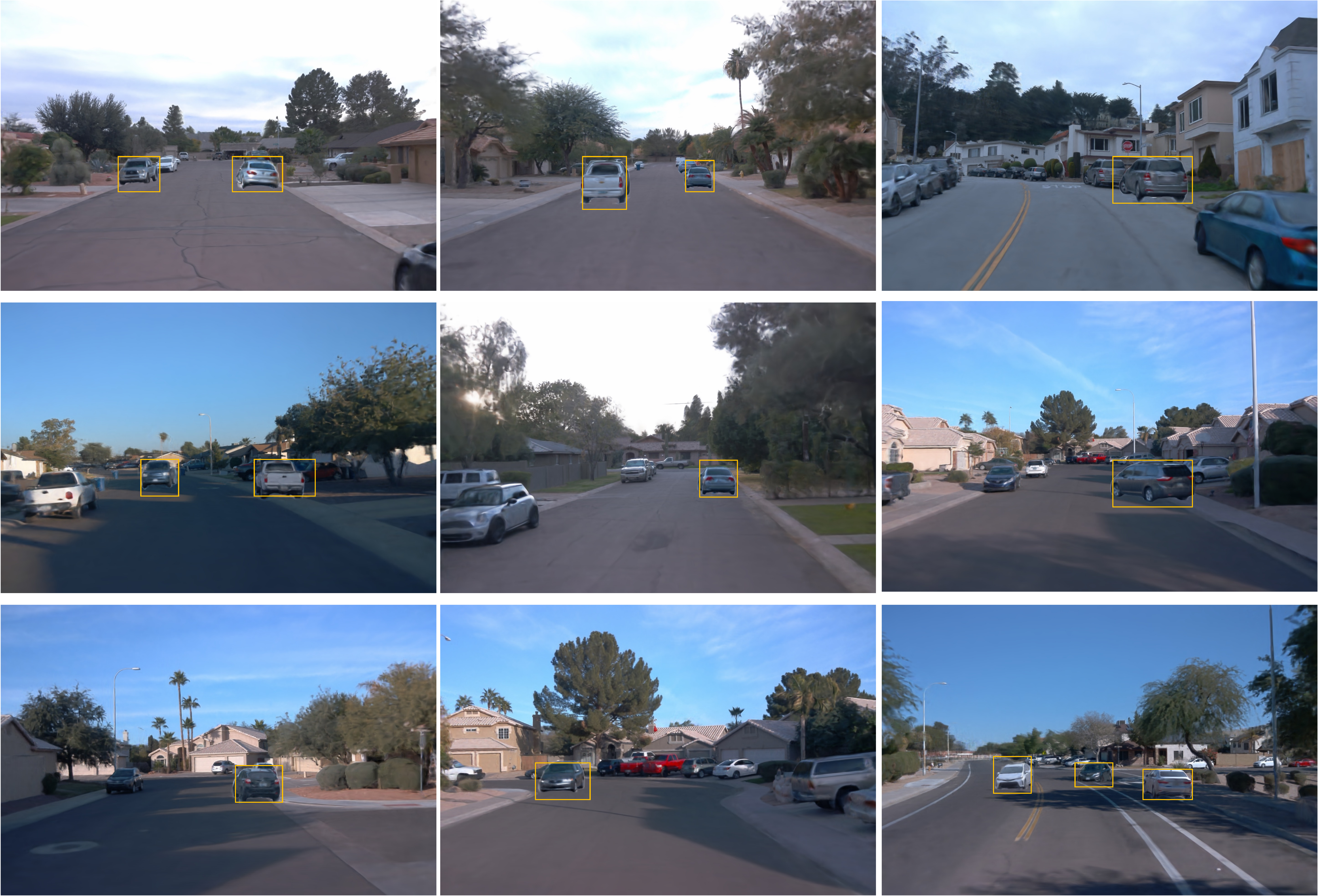}
    \end{center}
    \caption{\textbf{Visualization of the generated images by Drive-3DAug.} The yellow boxes indicate the newly added cars for the background.}
    \label{fig:more_vis}
\end{figure*}

\begin{figure*}[t]
    \begin{center}
        \includegraphics[width=0.9\linewidth]{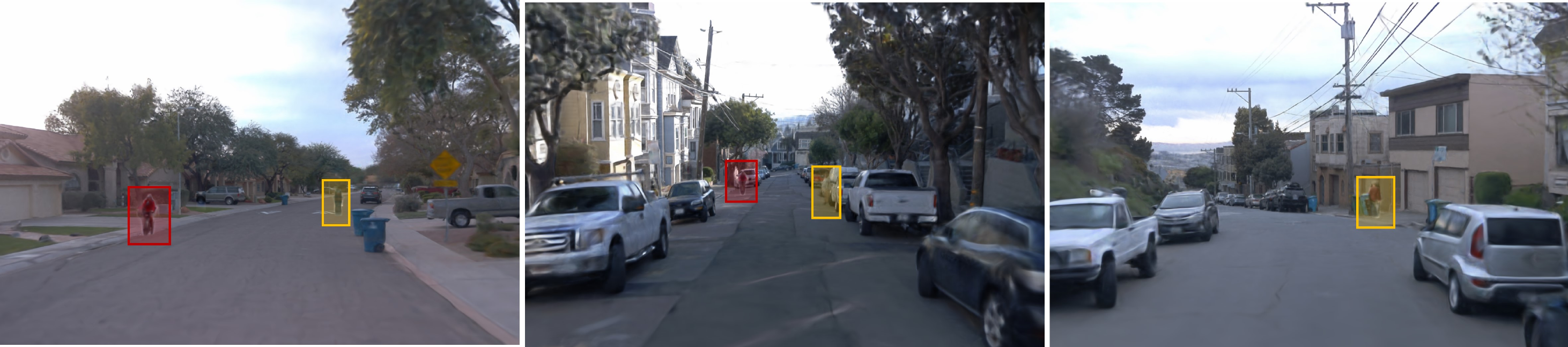}
    \end{center}
    \caption{\textbf{Visualization of the generated images by Drive-3DAug.} The yellow and red boxes indicate the augmented pedestrian and cyclist, respectively.}
    \label{fig:person}
\end{figure*}

We augment car in Figure~\ref{fig:more_vis}, indicating that
we can generate scene with high quality with Drive-3DAug.
Compared with car, pedestrian and cyclist are not rigid body and the size is small, not well applicable for NeRF modelling.
We model pedestrian and cyclist as rigid boby in the present study, which can cause the decay of object model performance.
As shown in Figure~\ref{fig:person}, although there exists flaw for augmented 
pedestrian and cyclist, we can still augment them to improve the detector performance.

\section{Cross-dataset Drive-3DAug}
\begin{figure*}[t]
    \begin{center}
        \includegraphics[width=0.9\linewidth]{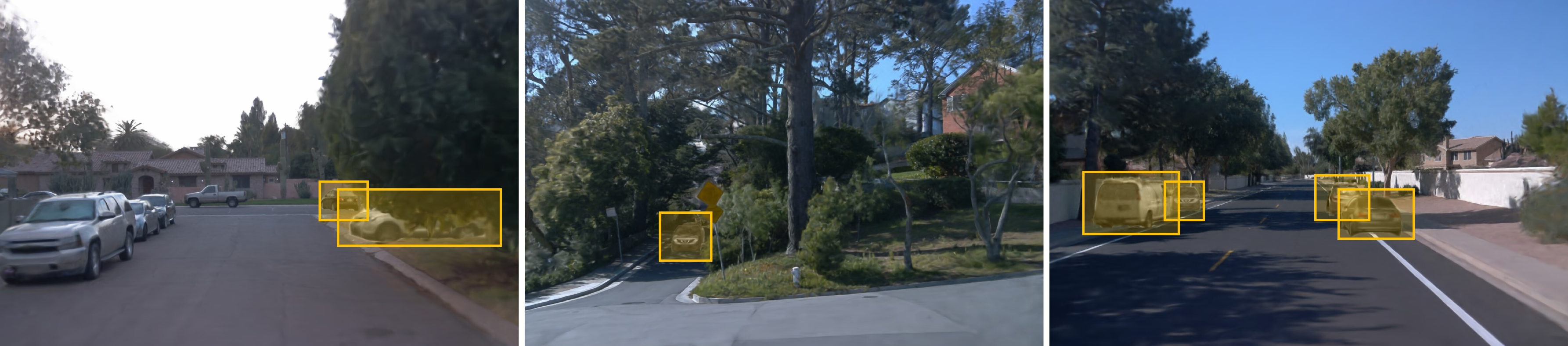}
    \end{center}
    \caption{\textbf{Image Generation cross datasets.} We place the cars from nuScenes on the backgrounds of Waymo. The yellow boxes indicate the augmented cars.}
    \label{fig:cross}
\end{figure*}

We have reconstructed thousands of background and object models in Waymo and nuScenes dataset.
These models can serve as the general model assets, convenient for creating new driving scenes inside a specific dataset or cross different datasets.
As shown in Figure \ref{fig:cross}, we compose the object models from nuScenes and the background models from Waymo to create new driving scenes. This can further enlarge the diversity of the training data, and we can generate large amounts of data for the study of model generalization across different datasets.

\end{document}